%% file: paper.tex
\documentclass[runningheads,a4paper]{llncs}

\usepackage{amssymb}
\usepackage{graphicx}

\usepackage{url}
\urldef{\mailsa}\path|{mina.rezaei, haojin.yang, christoph.meinel}@hpi.de
{konstantin.harmuth, willi.gierke, thomas.kellermeier, martin.fischer}@student.hpi.uni-potsdam.de|   
\newcommand{\keywords}[1]{\par\addvspace\baselineskip
\noindent\keywordname\enspace\ignorespaces#1}

\begin{document}

\mainmatter  

\title{Conditional Adversarial Network for Semantic Segmentation of Brain Tumor}

\titlerunning{cGANs for Semantic Segmentation of Brain Tumor}

%
%
\author{Mina Rezaei, Konstantin Harmuth, Willi Gierke, Thomas Kellermeier, Martin Fischer, Haojin Yang, Christoph Meinel}
\authorrunning{M.Rezaei et al.}

\institute{Hasso Plattner Institute for Digital Engineering,\\
Prof.-Dr.-Helmert-Straße 2-3, 14482 Potsdam, Germany\\
\mailsa\\ }

%
%
\toctitle{Lecture Notes in Computer Science}
\tocauthor{Authors' Instructions}
\maketitle

\input{chapters/abstract.tex}

\input{chapters/introduction.tex}

\input{chapters/methodology.tex}

\input{chapters/experiment.tex}

\input{chapters/conclusion.tex}

\bibliographystyle{splncs03}
\bibliography{reference}

\end{document}

%% file: chapters/abstract.tex
\begin{abstract}
Automated medical image analysis has a significant value in diagnosis and treatment of lesions.
Brain tumors segmentation has a special importance and difficulty due to the difference in appearances and shapes of the different tumor regions in magnetic resonance images.
Additionally the data sets are heterogeneous and usually limited in size in comparison with the computer vision problems.
The recently proposed adversarial training has shown promising results in generative image modeling.
In this paper we propose a novel end-to-end trainable architecture for brain tumor semantic segmentation through conditional adversarial training.
We exploit conditional Generative Adversarial Network (cGAN) and train a semantic segmentation Convolution Neural Network (CNN) along with an adversarial network that discriminates segmentation maps coming from the ground truth or from the segmentation network for BraTS 2017 segmentation task\cite{Menze2014,Bakasnature2017,Bakastcg2017,Bakaslgg2017}.
We also propose an end-to-end trainable CNN for survival day prediction based on deep learning techniques for BraTS 2017 prediction task~\cite{Menze2014,Bakasnature2017,Bakastcg2017,Bakaslgg2017}.
The experimental results demonstrate the superior ability of the proposed approach for both tasks.
The proposed model achieves on validation data a DICE score, Sensitivity and Specificity respectively 0.68, 0.99 and 0.98 for the whole tumor, regarding online judgment system.
\keywords{Conditional Generative Adversarial Network, Brain Tumor Semantic Segmentation, Survival day prediction}
\end{abstract}

%% file: chapters/introduction.tex
\section{Introduction}

Medical imaging plays an important role in disease diagnosis and treatment planning as well as clinical monitoring.
The diversity of magnetic resonance imaging (MRI) acquisition regarding its settings
(e.g. echo time, repetition time, etc.) and geometry (2D vs. 3D)
also the difference in hardware (e.g. field strength, gradient performance, etc.)
can yield variation in the appearance of the tumors that makes the automated segmentation challenging~\cite{cancers6010226}. An accurate brain lesion segmentation algorithm based on multi-modal MR images might be able to improve the prediction accuracy and efficiency for a better treatment planning and monitoring the disease progress.
As mentioned by Menze et al.~\cite{Menze2014}, in last few decades the number of clinical study for automatic brain lesion detection has grown significantly.
In the last three years, Generative Adversarial Network(GAN)~\cite{2014arXiv1406} become a very popular approach in various computer vision studies for example for classification~\cite{NIPS2016_6111,MakhzaniSJG15}, object detection~\cite{2017arXiv170605274L,wang2017fast}, video prediction \cite{MathieuCL15,FinnGL16,VondrickPT16}, image segmentation\cite{Phillipimagetoimage2017} and even mass segmentation for mammogram analysis~\cite{WentaoZhuX16}.
In this work we address two tasks by BraTS-2017~\cite{Menze2014,Bakasnature2017,Bakastcg2017,Bakaslgg2017} challenges by two different approaches.
Semantic segmentation is the task of classifying parts of images together that belong to the same object class.
Inspired by the power of cGAN networks~\cite{WentaoZhuX16,Phillipimagetoimage2017}, we propose an end-to-end trained adversarial deep structural network to perform brain High and Low Grade Glioma (HGG/LGG) tumor segmentation.
We also illustrate how this model could be used to learn a multi-modal images,
and provide preliminary results of an application for semantic segmentation.
To this end we consider patient-wise "U-Net"~\cite{ronneberger2015u} as a generator and "Markovian GAN"~\cite{LiW16b} as an discriminator.
For the second task of BraTS-2017~\cite{Menze2014,Bakasnature2017,Bakastcg2017,Bakaslgg2017}, we designed an end-to-end trainable CNNs on clinical data which enables to predict the survival day.
The architecture use parallel CNN which one way is responsible to learn patient-wise MR images and another learned representation of clinical data.
A detailed evaluation of the parameters variations and network architecture is provided.
The contribution of this work can be summarized as following:

\begin{itemize}
\item We proposed a robust solution for brain tumors segmentation through conditional GAN.
We achieved promising results on two type of brain tumor segmentation (The overall Dice for whole-tumor  region is 0.68, Specificity 0.99 and Sensitivity 0.98).
\item We proposed an automatic and trainable deep learning architecture for survival day prediction based on clinical data and MR images.
\end{itemize}

The rest of the paper is organized as follows: Chapter~\ref{methodology} describes the proposed approaches for semantic segmentation and survival day prediction, Chapter~\ref{experiment} presents the detailed experimental results.
Chapter~\ref{conclusion} concludes the paper and gives an outlook on future work.

%% file: chapters/methodology.tex
\section{Methodology} \label{methodology}

\begin{figure} [!t]
\includegraphics[width=1.0\textwidth]{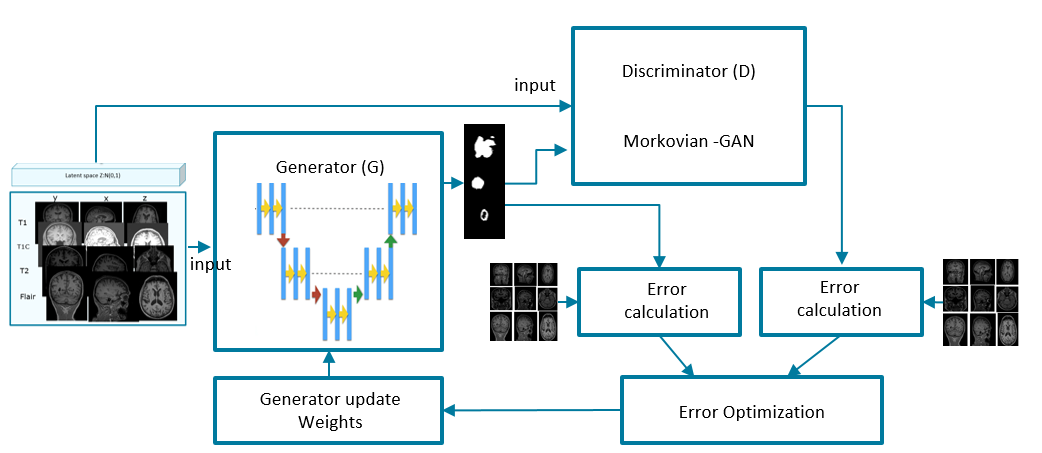}
\centering
\caption{The proposed architecture for semantic segmentation of brain tumor}
\label{fig_seg_arch}
\end{figure}

In this chapter we will describe first our proposed approach to the brain tumor sub-region segmentation based on deep learning and then our approach to the survival day prediction. 
The core techniques applied in our approach are depicted as well.
In the GAN theory~\cite{2014arXiv1406}, the Discriminator Network (D) tries to decide if a certain input is sourced from the reference distribution, or has been generated by the Generator Network (G).
The training procedure in G uses the pixel labels of certain multi-modal images and D tries to distinguish this certain boundary regions (we have three sub region tumor) comes from reference distribution or generative network.
In order to incorporate more classes to this output while keeping with the GAN spirit of distinguishing distribution class instead of one example class, we could add additional input sources.
As suggested by Goodfellow~\cite{2014arXiv1406}, one can consider the cGAN models with multi-class labels as:

\begin{enumerate}
  \item GAN model with class-conditional models: which make the input label rather than the output. We ask GAN to generate specic classes.~\cite{MirzaO14}
  \item GAN model with N different output classes: that network trained by N different "real" and no "fake" classes.~\cite{2015arXiv151106390S}
  \item GAN models with N+1 different output classes: which the network train by N different "real" and an additional "fake" class. This type works very well for semi-supervised learning when it combined with feature matching GANs e.g.~\cite{SalimansGZCRC16}
\end{enumerate}
Therefor our proposed method lies in the second category as we consider for each multi-modal image three segmentation classes.
Figure~\ref{fig_seg_arch} describes the proposed approach to the brain tumor segmentation.
In continue we describe the detail of techniques of pixel label classes for prediction in section~\ref{task1} and for survival day prediction in section~\ref{task2}.

\subsection{Brain Tumor Semantic Segmentation} \label{task1}

We adapt the generator and discriminator architectures from~\cite{RadfordMC15,Phillipimagetoimage2017}.
We applied Virtual-BatchNorm-Convolution~\cite{Goodfellow17tutor} on generator network to make the "U-Net"~\cite{ronneberger2015u} patient-wise.
We choose "U-Net" architecture as generator because most of the deep learning approaches are patch-wise learning models, which ignore the contextual information within the whole image region.
Like winner of BraTS-2016~\cite{urlmici}, we come over this problem by leveraging global-based CNN methods (e.g. Seg-Net, Encoder-Decoder and FCN) and incorporating multi-modal of MRI data.
We use Virtual-BatchNorm~\cite{Goodfellow17tutor} in the generator network and Reference-BatchNorm~\cite{Goodfellow17tutor}in the discriminator network to reduce over-fitting.
The discriminative network is based on "Markovian GAN"~\cite{RadfordMC15}.
Then two models trainable simultaneously through back propagation, corresponds to a minimax two-player game.
An "U-Net" generative model G; Captures the data distribution, pixel segmentation and train to minimize the probability of D making a mistake.
A "Markovian GAN" discriminative model D: to estimate the probability that a sample came from the training data rather than G.

\subsection{Survival Day Prediction} \label{task2}

\begin{figure}
\includegraphics[width=0.9\textwidth]{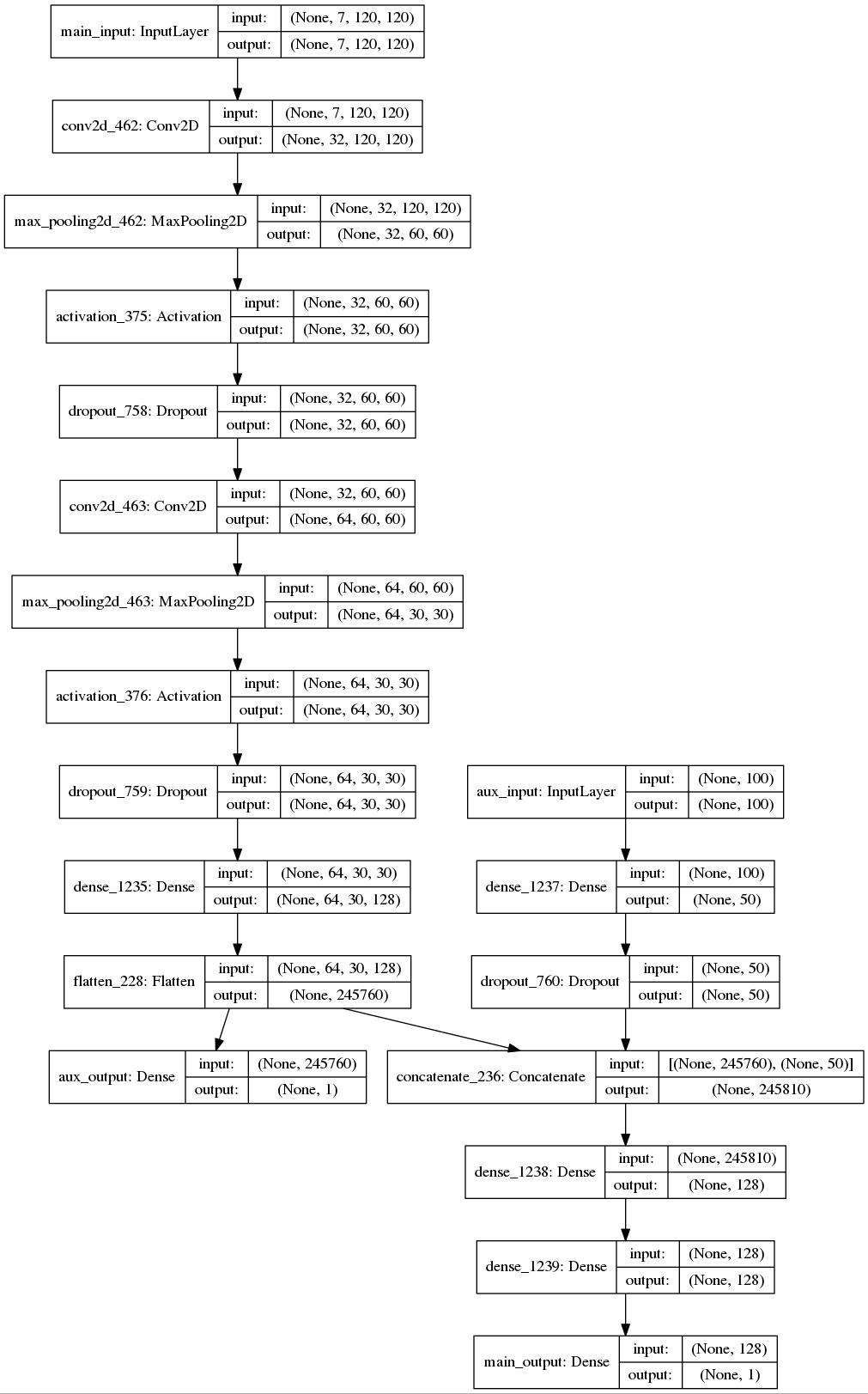}
\centering
\caption{The propose architecture for survival day prediction}
\label{fig_sv_arch}
\end{figure}

Figure~\ref{fig_sv_arch} describes our solution for survival day prediction.
We proposed a two path way architecture which one has several CNN and it is responsible for multi-modal image representation and another learned the clinical data features.
The extracted features from each path way, concatenated in next step to shared the learned features.
Then they passed to two fully connected layers to learn the survival day.
We use Virtual-BatchNorm~\cite{Goodfellow17tutor} on the CNNs network which learned image representation.
To prevent over-fitting, we generated augmented images through horizontal and vertical flipping and re-scaling.
We applied Mean squared error as Loss function.
We mapped the clinical data (Ages and survival days) into float[0,1].

%% file: chapters/experiment.tex
\section{Experiments}  \label{experiment}

\begin{figure}
\includegraphics[width=0.9\textwidth]{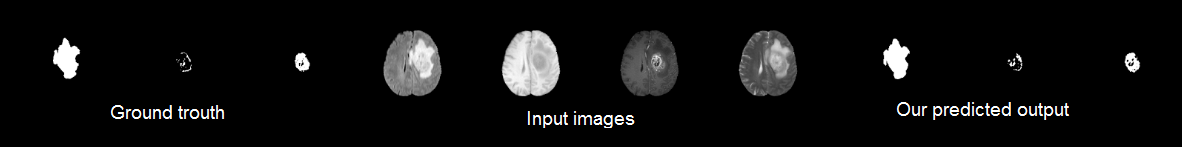}
\centering
\caption{The output segmentation result on training data}
\label{fig_segtrain}
\end{figure}

In order to evaluate the performance of the proposed cGANs method, we test the method on two types of brain tumor data provided by BraTs 2017 challenge~\cite{Menze2014,Bakasnature2017,Bakastcg2017,Bakaslgg2017}.
We applied a bias field correction on the MR images to correct the intensity non-uniformity in MR images by using N4ITK~\cite{tustison2010n4itk}.
In next step of pre-processing we applied histogram matching normalization~\cite{Jayaram200}.
We train both the generator and the discriminator to make them stronger together and avoid making one network significantly stronger than the other by taking turn.
We consider multi-madal images from same patient in each batch during training and use all the released data by BraTS 2017 challenge\cite{Menze2014,Bakasnature2017,Bakastcg2017,Bakaslgg2017} in training time which is 75 patients with Low Grade Glioma(LGG) and 210 patients with High Grade Glioma(HGG).
We used all prepared image-modal from three axes of x,y,z (3x4x155x285) that the input and output are 4-3 channel images(4:image-modal; 3:three sub-region of each tumor type).
We get better result when don't shuffle input data in generator network.
In generator network Sign function helps for noise reduction.
The generator for all layers use ReLU activation function except output layer which use Tanh. 
Qualitative results are shown in Figures~\ref{fig_segtrain}. 
On this size data sets (530100 2D images with the size of 250x250) training took around 72 hours on parallel Pascal Titan X GPUs.
Table 1\ref{table-validation-result} shows the results of the proposed models evaluated at BraTS 2017 online judge system.
The evaluation system uses three tasks.
The online system provides the results as follows:
The tumor structures are grouped in three different tumor regions.
This is mainly due to practical clinical applications.
As described by BraTS 2017~\cite{Menze2014,Bakasnature2017,Bakastcg2017,Bakaslgg2017}, tumor regions are defined as: 
\begin{enumerate}
  \item WT: Whole tumor region represents the area with all labels 1,2,3,4 which 0 for normal tissue, 1 for edema, 2 for non-enhancing core, 3 for necrotic core, 4 shows enhancing core.
  \item CT: Core tumor region represent only tumor core region, it measures label 1,3,4.
  \item ET: Enhancing tumor region (label 4)
\end{enumerate}

There are four kinds of evaluation criteria for segmentation task like Dice score, Hausdorff distance, Sensitivity and Specificity has provided by BraTS 2017 challenge organizer as an online judgment system.

\begin{table}[]
\centering
\caption{Preliminary results till now from BraTS-2017 online judge system on Validation data(unseen data)}
\label{table-validation-result}
\begin{tabular}{|l|l|l|l|l|l|l|}
\hline
        ~ & Whole Tumor & Core of Tumor & Enhanced Tumor \\
\hline
        Dice & 0.70 & 0.55 & 0.40  \\ 
        Sensitivity & 0.68 & 0.52 & 0.99 \\
        Specificity & 0.99 & 0.99 & 0.99 \\
\hline
\end{tabular}
\end{table}

\begin{figure}
\includegraphics[width=0.9\textwidth]{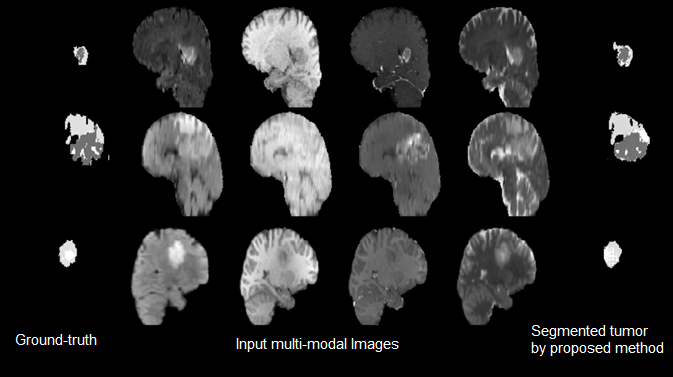}
\centering
\caption{The output segmentation result on training data}
\label{fig_segtrain3region}
\end{figure}

\begin{figure}
\includegraphics[width=0.9\textwidth]{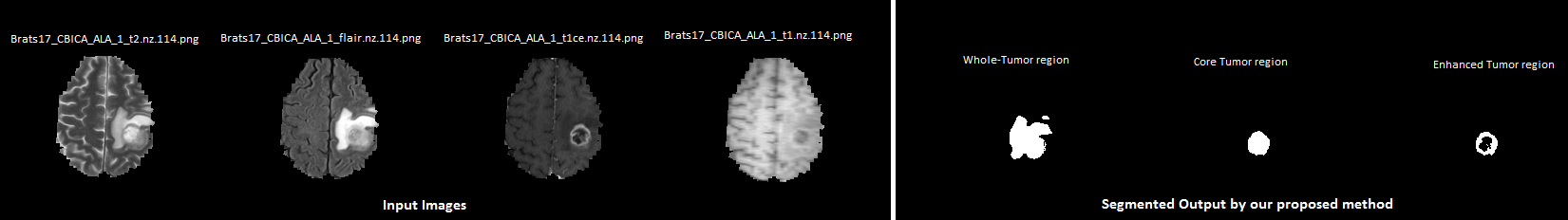}
\centering
\caption{The preliminary segmentation result on validation data}
\label{fig_segval}
\end{figure}

Table 1 shows the preliminary results but our work is still on the progress. Table\ref{table_sv_result} shows the survival day prediction results.

\begin{table}[]
\centering
\caption{Preliminary results on survival day prediction. We used 70\% of the data (115 patients) for training, 10\% (16 patients) for validation and 20\% (32 patients) for
testing. The first path way of CNN has seven input channel which four from multi-modal images and three from segmented regions. We translated ages from interval [0, 100] into float [0,1] and also for survival day did from [0-1750] days into float of [0,1].}
\label{table_sv_result}
\begin{tabular}{|l|l|l|}
\hline
        Data & Accuracy  \\
\hline
        Validation & 73.1\%  \\ 
        Test & 64.08\%  \\

\hline
\end{tabular}
\end{table}

\begin{figure}
\includegraphics[width=0.9\textwidth]{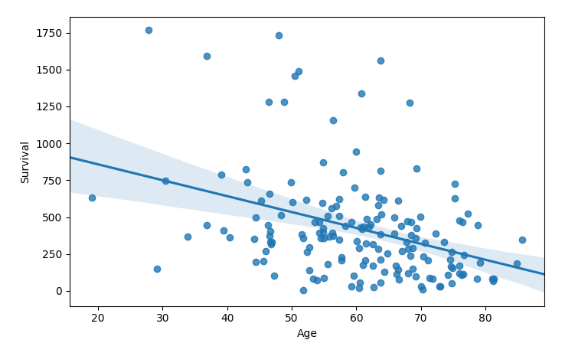}
\centering
\caption{clinical data distribution from training set}
\label{fig_svdis}
\end{figure}

\begin{figure}
\includegraphics[width=0.9\textwidth]{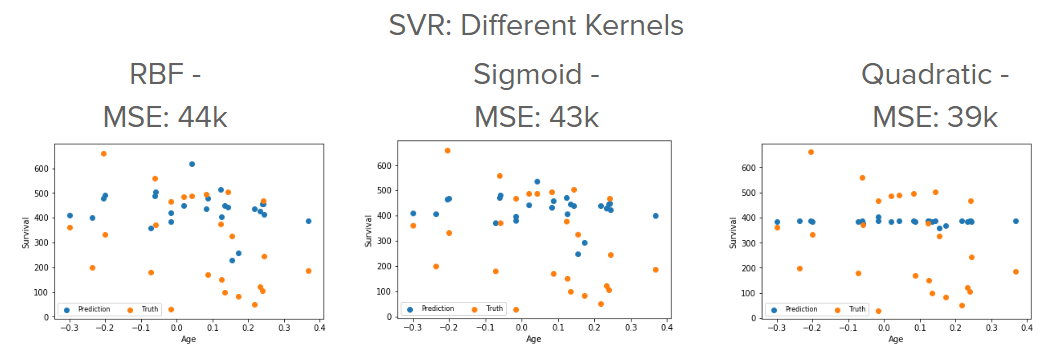}
\centering
\caption{different regression techniques (e. g. Support Vector Regression, Polynomial Regression, …) for survival day prediction.}
\label{fig_svrk}
\end{figure}

%% file: chapters/conclusion.tex
\section{Conclusion} \label{conclusion}

In this paper, we propose and evaluated approaches for two important clinical tasks: brain tumor segmentation and prediction of survival day after tumor diagnosis.
The proposed approach for tumor segmentation is end-to-end trainable based on the newly proposed conditional generative adversarial network.
Furthermore, adversarial training is used to handle the global-based CNN in generator to reduce over-fitting and increase robustness.
We proposed an automated trainable parallel convolution neural network to predict the survival day as the second task in the challenge.
These networks learn a loss adapted to the task and data at hand, which makes it applicable in unseen data.
For the future work, we look for further improvement on generative network by incorporating recurrent neural network(RNN) inside of our Encoder-Decoder.

%% file: paper.bbl
\begin{thebibliography}{10}
\providecommand{\url}[1]{\texttt{#1}}
\providecommand{\urlprefix}{URL }

\bibitem{urlmici}
 \url{https://www.cbica.upenn.edu/sbia/Spyridon.Bakas/MICCAI_BraTS/MICCAI_BraTS_2016_proceedings.pdf}

\bibitem{Bakastcg2017}
{Bakas}, S., {Akbari}, H., {Sotiras}, A., {Bilello}, M., {Rozycki}, M.,
  {Kirby}, J., {Freymann}, J., {Farahani}, K., {Davatzikos}, C.: {Segmentation
  Labels and Radiomic Features for the Pre-operative Scans of the TCGA-GBM
  collection}. The Cancer Imaging Archive  (2017)

\bibitem{Bakaslgg2017}
{Bakas}, S., {Akbari}, H., {Sotiras}, A., {Bilello}, M., {Rozycki}, M.,
  {Kirby}, J., {Freymann}, J., {Farahani}, K., {Davatzikos}, C.: {Segmentation
  Labels and Radiomic Features for the Pre-operative Scans of the TCGA-LGG
  collection}. The Cancer Imaging Archive  (2017)

\bibitem{Bakasnature2017}
{Bakas}, S., {Akbari}, H., {Sotiras}, A., {Bilello}, M., {Rozycki}, M.,
  {Kirby}, J., {Freymann}, J., {Farahani}, K., {Davatzikos}, C.: {Advancing The
  Cancer Genome Atlas glioma MRI collections with expert segmentation labels
  and radiomic features}. Nature Scientific Data  (2017)

\bibitem{FinnGL16}
Finn, C., Goodfellow, I.J., Levine, S.: Unsupervised learning for physical
  interaction through video prediction. CoRR  abs/1605.07157 (2016),
  \url{http://arxiv.org/abs/1605.07157}

\bibitem{2014arXiv1406}
{Goodfellow}, I.J., {Pouget-Abadie}, J., {Mirza}, M., {Xu}, B., {Warde-Farley},
  D., {Ozair}, S., {Courville}, A., {Bengio}, Y.: {Generative Adversarial
  Networks}. ArXiv e-prints  (2014)

\bibitem{Goodfellow17tutor}
Goodfellow, I.J.: {NIPS} 2016 tutorial: Generative adversarial networks. CoRR
  abs/1701.00160 (2017)

\bibitem{cancers6010226}
Inda, Maria-del-Mar, R.B., Seoane, J.: Glioblastoma multiforme: A look inside
  its heterogeneous nature. In: Cancer Archive 226-239 (2014)

\bibitem{Phillipimagetoimage2017}
Isola, P., Zhu, J., Zhou, T., Efros, A.A.: Image-to-image translation with
  conditional adversarial networks. CoRR  abs/1611.07004 (2016),
  \url{http://arxiv.org/abs/1611.07004}

\bibitem{LiW16b}
Li, C., Wand, M.: Precomputed real-time texture synthesis with markovian
  generative adversarial networks. CoRR  abs/1604.04382 (2016),
  \url{http://arxiv.org/abs/1604.04382}

\bibitem{2017arXiv170605274L}
{Li}, J., {Liang}, X., {Wei}, Y., {Xu}, T., {Feng}, J., {Yan}, S.: {Perceptual
  Generative Adversarial Networks for Small Object Detection}. ArXiv e-prints
  (Jun 2017)

\bibitem{Jayaram200}
L�szl� G.~Ny�l, J.K.U., Zhang, X.: New variants of a method of mri scale
  standardization. IEEE transactions on medical imaging  29(6) (2000)

\bibitem{MakhzaniSJG15}
Makhzani, A., Shlens, J., Jaitly, N., Goodfellow, I.J.: Adversarial
  autoencoders. CoRR  abs/1511.05644 (2015),
  \url{http://arxiv.org/abs/1511.05644}

\bibitem{MathieuCL15}
Mathieu, M., Couprie, C., LeCun, Y.: Deep multi-scale video prediction beyond
  mean square error. CoRR  abs/1511.05440 (2015),
  \url{http://arxiv.org/abs/1511.05440}

\bibitem{Menze2014}
{Menze}, B., {Jakab}, A., {Bauer}, S., { Kalpathy-Cramer}, J., {Farahani}, K.,
  {Kirby}, J., {Burren}, Y., {Porz}, N., {Slotboom}, J., {Wiest}, R., {Lanczi},
  L., {Gerstner}, E., {Weber}, M., {Arbel}, T., {Avants}, B., {Ayache}, N.,
  {Buendia}, P., {Collins}, D., {Cordier}, N., {Corso}, J., {Criminisi}, A.,
  {Das}, T., {Delingette}, H., {Demiralp}, ., {Durst}, C., {Dojat}, M.,
  {Doyle}, S., {Festa}, J., {Forbes}, F., {Geremia}, E., {Glocker}, B.,
  {Golland}, P., {Guo}, D., {Hamamci}, A., {Iftekharuddin}, K., {Jena}, R.,
  {John}, N., {Konukoglu}, E., {Lashkari}, D., {Mariz}, J., {Meier}, R.,
  {Pereira}, S., {Precup}, D., {Price}, S., {Raviv}, T., {Reza}, S., {Ryan},
  S., {Sarikaya}, D., {Schwartz}, L., {Shin}, H., {Shotton}, J., {Silva}, C.,
  {Sousa}, N., {Subbanna}, N., {Szekely}, G., {Taylor}, T., {Thomas}, O.,
  {Tustison}, N., {Unal}, G., {Vasseur}, F., {Wintermark}, M., {Ye}, D.,
  {Zhao}, L., {Zhao}, B., {Zikic}, D., {Prastawa}, M., {Reyes}, M., {Van
  Leemput}, K.: {The multimodal brain tumor image segmentation benchmark
  (BRATS)}. IEEE transactions on medical imaging  34(10),  1993--2024 (2015)

\bibitem{MirzaO14}
Mirza, M., Osindero, S.: Conditional generative adversarial nets. CoRR
  abs/1411.1784 (2014), \url{http://arxiv.org/abs/1411.1784}

\bibitem{RadfordMC15}
Radford, A., Metz, L., Chintala, S.: Unsupervised representation learning with
  deep convolutional generative adversarial networks. CoRR  abs/1511.06434
  (2015), \url{http://arxiv.org/abs/1511.06434}

\bibitem{NIPS2016_6111}
Reed, S.E., Akata, Z., Mohan, S., Tenka, S., Schiele, B., Lee, H.: Learning
  what and where to draw. In: Lee, D.D., Sugiyama, M., Luxburg, U.V., Guyon,
  I., Garnett, R. (eds.) Advances in Neural Information Processing Systems 29,
  pp. 217--225. Curran Associates, Inc. (2016),
  \url{http://papers.nips.cc/paper/6111-learning-what-and-where-to-draw.pdf}

\bibitem{ronneberger2015u}
Ronneberger, O., Fischer, P., Brox, T.: U-net: Convolutional networks for
  biomedical image segmentation. In: International Conference on Medical Image
  Computing and Computer-Assisted Intervention. pp. 234--241. Springer
  International Publishing (2015)

\bibitem{SalimansGZCRC16}
Salimans, T., Goodfellow, I.J., Zaremba, W., Cheung, V., Radford, A., Chen, X.:
  Improved techniques for training gans. CoRR  abs/1606.03498 (2016),
  \url{http://arxiv.org/abs/1606.03498}

\bibitem{2015arXiv151106390S}
{Springenberg}, J.T.: {Unsupervised and Semi-supervised Learning with
  Categorical Generative Adversarial Networks}. ArXiv e-prints  (Nov 2015)

\bibitem{tustison2010n4itk}
Tustison, N.J., Avants, B.B., Cook, P.A., Zheng, Y., Egan, A., Yushkevich,
  P.A., Gee, J.C.: N4itk: improved n3 bias correction. IEEE transactions on
  medical imaging  29(6),  1310--1320 (2010)

\bibitem{VondrickPT16}
Vondrick, C., Pirsiavash, H., Torralba, A.: Generating videos with scene
  dynamics. CoRR  abs/1609.02612 (2016), \url{http://arxiv.org/abs/1609.02612}

\bibitem{wang2017fast}
Wang, X., Shrivastava, A., Gupta, A.: A-fast-rcnn: Hard positive generation via
  adversary for object detection. arXiv preprint arXiv:1704.03414  (2017)

\bibitem{WentaoZhuX16}
Zhu, W., Xie, X.: Adversarial deep structural networks for mammographic mass
  segmentation. CoRR  abs/1612.05970 (2016),
  \url{http://arxiv.org/abs/1612.05970}

\end{thebibliography}
